\documentclass[10pt,twocolumn,letterpaper]{article}

\usepackage[pagenumbers]{cvpr} 

\usepackage{multirow}
\usepackage{colortbl}
\usepackage{makecell}
\usepackage{algorithm}
\usepackage{algpseudocode}
\usepackage{bm}
\usepackage{amsthm}
\usepackage{xspace}
\usepackage{microtype}

\newcommand{\method}{ExiST\xspace}
\newcommand{\methodlong}{Existence-aware Self-Training\xspace}

\definecolor{first}{RGB}{242,170,170}
\definecolor{second}{RGB}{247,205,160}
\definecolor{third}{RGB}{252,240,170}

\DeclareMathOperator*{\argmax}{arg\,max}
\newcommand{\R}{\mathbb{R}}

\newcommand{\ind}[1]{\mathbb{1}\!\left[#1\right]}
\newcommand{\Dl}{\mathcal{D}_{L}}
\newcommand{\Du}{\mathcal{D}_{U}}
\newcommand{\Dv}{\mathcal{D}_{V}}
\newcommand{\Lsup}{\mathcal{L}_{\mathrm{sup}}}
\newcommand{\Lexist}{\mathcal{L}_{\mathrm{exist}}}
\newcommand{\Lmce}{\mathcal{L}_{\mathrm{mCE}}}
\newcommand{\Ldice}{\mathcal{L}_{\mathrm{Dice}}}
\newcommand{\Lwce}{\mathcal{L}_{\mathrm{wCE}}}
\newcommand{\cS}{\mathcal{S}}
\newcommand{\cA}{\mathcal{A}}
\newcommand{\cB}{\mathcal{B}}
\newcommand{\cU}{\mathcal{U}}
\newcommand{\cK}{\mathcal{K}}
\newcommand{\cC}{\mathcal{C}}
\newcommand{\Om}{\Omega}

\definecolor{cvprblue}{rgb}{0.21,0.49,0.74}
\usepackage[pagebackref,breaklinks,colorlinks,allcolors=cvprblue]{hyperref}
\usepackage[capitalize,nameinlink]{cleveref}
\crefname{section}{Sec.}{Secs.}
\Crefname{section}{Section}{Sections}
\crefname{table}{Tab.}{Tabs.}
\Crefname{table}{Table}{Tables}
\crefname{figure}{Fig.}{Figs.}
\Crefname{figure}{Figure}{Figures}
\crefname{equation}{Eq.}{Eqs.}
\Crefname{equation}{Equation}{Equations}
\crefname{algorithm}{Alg.}{Algs.}
\crefname{proposition}{Prop.}{Props.}
\Crefname{proposition}{Proposition}{Propositions}

\newtheorem{proposition}{Proposition}

\def\confName{CVPR}
\def\confYear{2027}

\title{ExiST: Existence-Aware Self-Training for\\Semi-Supervised Multi-Organ Segmentation}

\author{
Maxx Richard Rahman$^{1,2}$ \and
Prakhar Kumar$^{1}$ \and
Wolfgang Maass$^{1,2}$ \\
\\
$^{1}$Saarland University, Germany \\
$^{2}$German Research Center for Artificial Intelligence (DFKI), Germany
}

\begin{document}
\maketitle
\begin{abstract}
Semi-supervised multi-organ segmentation should learn more than a dozen organs from a handful of labeled scans, and the smallest organs occupy less than one percent of the foreground. We identify a failure mode of self-training that is specific to this regime. When a teacher misses a small organ in a scan, every voxel of that organ receives the confident pseudo-label background, and the student is trained toward the organ's absence. Repeated over many scans, this loop extinguishes tail organs. In a pilot study, removing the gallbladder from the pseudo-labels reduced its Dice on AMOS from 36.5 to 1.0. We argue that a missing organ should be treated as unresolved rather than absent and propose \methodlong (\method). It builds pseudo-labels by leave-one-out consensus of independently trained networks, screens them with a quality gate on validation scans, and infers, per scan and per organ, whether the organ was found. For every unresolved organ, the background voxels of that scan receive the set-valued target, trained with a marginal cross-entropy that accepts either answer and a Dice term that skips uncertain voxels. We show that this loss never pushes a candidate organ toward absence and never favors one member of the set, which leaves the decision to labeled evidence. \method outperforms the current state-of-the-art methods and reaches 77.1 mean Dice on Synapse with 20\% labeled scans and 75.9 on AMOS with 5\% labeled scans.
\end{abstract}

\section{Introduction}
\label{sec:intro}

\begin{figure}[t]
  \centering
  \includegraphics[width=\linewidth]{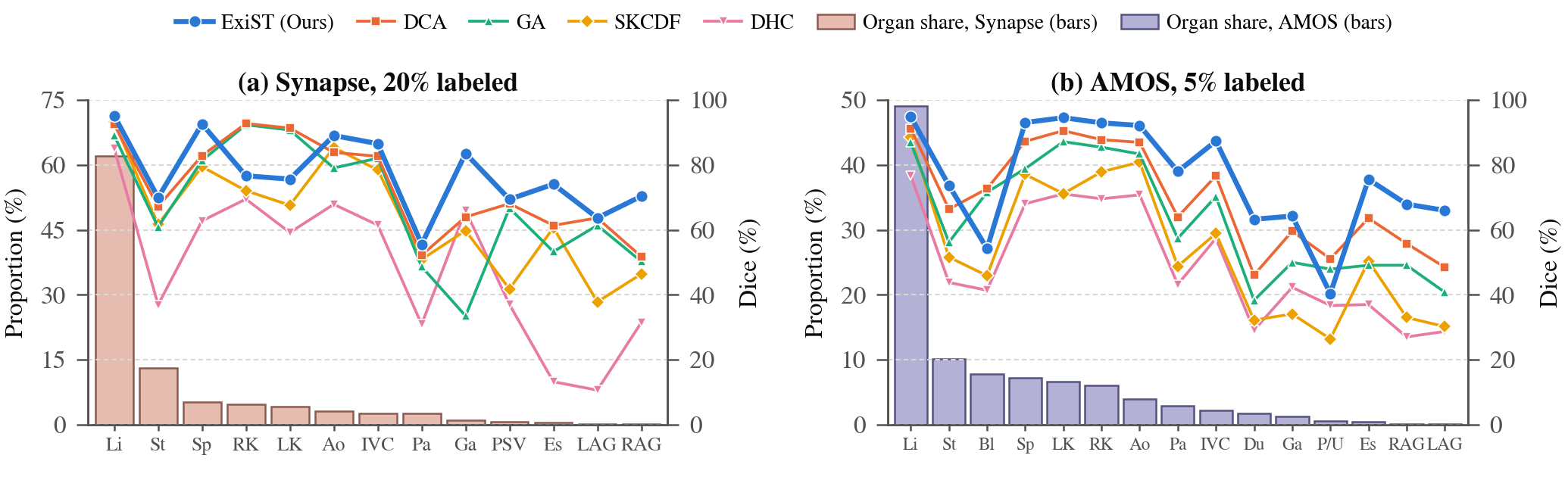}
  \caption{Per-organ Dice (lines) on (a) Synapse (20\% labeled) and (b) AMOS (5\% labeled). Organs are ordered by their share of foreground voxels in the labeled scans (bars).}
  \label{fig:per_organ}
\end{figure}

Multi-organ segmentation of abdominal CT and MRI underlies radiotherapy planning and surgical navigation, and voxel-level annotation of a dozen organs per scan is among the most expensive labeling tasks in medical imaging~\cite{ji2022amos,landman2015btcv}. Semi-supervised segmentation reduces this cost by training on a few labeled scans together with many unlabeled ones~\cite{tarvainen2017meanteacher,yu2019uamt,chen2021cps,bai2023bcp}, most often by self-training: a teacher, typically an exponential moving average (EMA) of the student or a co-trained sibling, predicts the unlabeled scans and the student reproduces its confident predictions~\cite{sohn2020fixmatch,yang2023unimatch}. Multi-organ data make this regime unusually hard. The liver holds roughly half of the foreground voxels of the labeled scans of Synapse and AMOS, whereas the adrenal glands and the esophagus each hold less than one percent (\cref{fig:per_organ}). Pseudo-labels inherit and amplify this imbalance, and general methods such as UA-MT~\cite{yu2019uamt} and CPS~\cite{chen2021cps} leave several tail organs at zero Dice (\cref{tab:synapse}). Recent work debiases pseudo-label confidence, re-weights classes, balances heads, or routes organs to experts~\cite{wang2023dhc,qi2024ga,zhang2025skcdf,liu2026dca}.

All of these methods share an assumption that we argue is a real bottleneck for tail organs: \emph{background is treated as certain}. When the teacher misses a small organ entirely, there is no organ pseudo-label to re-weight. The organ's voxels are confidently labeled background, the student is pushed toward the organ's absence, the EMA teacher inherits the weaker student, and the organ is missed in more scans. We call this loop \emph{pseudo-label extinction}. It is invisible to confidence filtering because the background labels that cause it are confident. A pilot study under the DHC protocol~\cite{wang2023dhc} shows the effect directly. When the gallbladder was removed from the pseudo-labels while its voxels stayed confident background, the mean test Dice of the student on AMOS fell from 37.4 to 29.7 and its gallbladder Dice from 36.5 to 1.0.

Our central insight is that the right response to a missed organ is not a better background label but a weaker one. If the pseudo-labels of a scan contain far less of organ $k$ than that anatomy normally contains, organ $k$ was not resolved, and the student should not be told where it is absent. We therefore replace the single background label by a \emph{set-valued target}. Every accepted background voxel of such a scan may be either background or organ $k$, and the student is trained with the marginal likelihood of the set. A truly background voxel is still fitted, while a voxel of the missed organ no longer generates a gradient toward absence. Set-valued targets are established when the set of annotated organs is known per dataset~\cite{shi2021marginal,zhou2019prior}. Here the set is inferred per scan and per organ from the pseudo-labeler itself. \method also decouples the pseudo-labeler from the student. Three networks are trained on the labeled scans with different seeds, each student is paired with the two networks that do not share its seed, a voxel is accepted only when both peers agree with high confidence, and a quality gate on the validation scans decides whether self-training may start.

\paragraph{Contributions:}
\begin{itemize}
  \item We identify pseudo-label extinction as a failure mode of self-training for multi-organ segmentation that confidence filtering and class re-weighting cannot fix, and formalize it through the gradient of background pseudo-labels on a missed organ.
  \item We propose existence-aware targets. A per-scan, per-organ test decides whether an organ was found, and unverified background is converted into set-valued targets trained with a marginal cross-entropy and an uncertainty-skipping Dice loss, which provably never decides within the set and never pushes a candidate organ toward absence. The targets are built from leave-one-out consensus pseudo-labels screened by a validation quality gate.
  \item \method sets a new state-of-the-art on Synapse and AMOS datasets, with the significant improvements on small organs and single-network inference.
\end{itemize}

\section{Related Work}
\label{sec:related}

\paragraph{Class-imbalanced semi-supervised medical image segmentation.}
General semi-supervised segmentation relies on EMA teachers~\cite{tarvainen2017meanteacher,yu2019uamt}, consistency constraints~\cite{luo2021urpc,wu2022ssnet}, mutual supervision~\cite{chen2021cps,peiris2023cobionet}, data mixing~\cite{bai2023bcp,chen2023magicnet,chi2024abd,hu2025betafft}, and contrastive or generative terms~\cite{assefa2025dycon,huang2026semigda}. These methods treat confident background as correct and degrade sharply on 13 to 15 organs. Class-imbalanced classification corrects the class prior of pseudo-labels~\cite{kim2020distribution,wei2021crest,guo2022adsh,oh2022daso,wang2022debiased,wei2023towards,zhang2021flexmatch,chen2023softmatch}, and for segmentation, CLD~\cite{lin2022cld} and Basak \etal~\cite{basak2022addressing} re-weight classes, DHC~\cite{wang2023dhc} co-trains heterogeneous networks with debiased weights, and A\&D~\cite{wang2023generic} aggregates and decouples heads. GA~\cite{qi2024ga} modifies the gradient of the Dice and cross-entropy losses so that classes absent from a label still receive a useful signal, SKCDF~\cite{zhang2025skcdf} decouples a balanced head, and DCA~\cite{liu2026dca} routes head, medium, and tail organs to asymmetric experts. All of these methods act on the organ side of the pseudo-labels, and none tests whether an organ was found in a scan. GA is closest in motivation, but it changes the loss on every voxel, whereas \method changes the target only where an organ is in doubt.

\paragraph{Set-valued targets and agreement-based pseudo-labels.}
Partially supervised segmentation faces a related problem. When datasets annotate different organs, the unannotated organs of a scan hide in its background. Prior-aware networks~\cite{zhou2019prior}, marginal and exclusion losses~\cite{shi2021marginal}, and later methods~\cite{fang2020multi,zhang2021dodnet,liu2023clip,liu2024cosst,jiang2025labeled,chen2024versatile} therefore train with a background class that marginalizes over the organs known to be unannotated. \method uses the same marginal likelihood, but the set of hidden organs is not dataset metadata. It is inferred per scan and per organ from the consensus of independent pseudo-labelers and applied to pseudo-labels rather than annotations. U2PL~\cite{wang2022semi} uses unreliable pixels as negatives for the classes they are confidently not, whereas \method withdraws the negative for the class the voxel may be. Learning from candidate label sets~\cite{jin2002learning,cour2011learning,feng2020provably,lv2020progressive,wang2022pico,chen2024imprecise} and negative learning~\cite{ishida2017learning,kim2019nlnl} provide the theory. Agreement-based pseudo-labels go back to co-training~\cite{blum1998combining,qiao2018deep,ke2020guided,ouali2020semi,xia2020uncertainty} and tri-training~\cite{zhou2005tri}, which label an example only when other learners agree. The leave-one-out consensus of \method follows tri-training, but it is applied once, offline, with frozen networks, and it yields set-valued targets. Foundation models~\cite{wasserthal2023totalsegmentator,ma2024segment,he2025vista3d,ma2025steady} could replace these peers.

\section{Preliminaries}
\label{sec:prelim}

\paragraph{Setting.}
A scan $x\in\R^{|\Om|}$ is defined on the voxel grid $\Om$ after preprocessing, with label map $y\in\cC^{|\Om|}$ over the classes $\cC=\{0,1,\dots,K\}$, where $0$ is background and $\cK=\{1,\dots,K\}$ indexes the organs ($K=13$ for Synapse, $15$ for AMOS). We are given a small labeled set $\Dl=\{(x_i,y_i)\}_{i=1}^{N_L}$, a large unlabeled set $\Du=\{x_i\}_{i=N_L+1}^{N_L+N_U}$, and a labeled validation set $\Dv$ used for model selection and for the quality gate of \cref{sec:gate}. A network $f_\theta$ maps a scan to per-voxel logits $z_\theta(v)\in\R^{K+1}$ and posteriors $p_\theta(c\mid v)=\mathrm{softmax}(z_\theta(v))_c$, with prediction $\hat y_\theta(v)=\argmax_c p_\theta(c\mid v)$. On labeled patches we minimize $\Lsup=\Lwce+\Ldice$, the sum of a class-weighted cross-entropy and a soft Dice loss that is standard for imbalanced 3D segmentation~\cite{isensee2021nnunet,wang2023dhc}. The weight $w_c$ of class $c$ in $\Lwce$ decreases with the voxel share of the class in $\Dl$.

\paragraph{Self-training and pseudo-label extinction.}
Self-training adds a loss on unlabeled scans whose targets come from a pseudo-labeler $g$, which can be an EMA teacher~\cite{tarvainen2017meanteacher}, a co-trained sibling~\cite{chen2021cps}, or, as here, a set of frozen networks. The usual target assigns one class per voxel and keeps only confident voxels,
\begin{equation}
  \hat y(v)=\argmax_c p_g(c\mid v),\quad m(v)=\ind{\max_c p_g(c\mid v)\ge\gamma},
  \label{eq:hard_pl}
\end{equation}
and the unlabeled loss is $\sum_v m(v)\,\ell\big(p_\theta(\cdot\mid v),\hat y(v)\big)$ with $\ell$ a cross-entropy. Class-imbalance methods modify $\gamma$, $w_c$, or the sampling of organ voxels, but they keep the structure of \cref{eq:hard_pl}: a voxel is either assigned one class or discarded. Consider an organ $k$ that $g$ fails to detect in scan $x_i$, so that $p_g(k\mid v)$ is small on the true voxels of the organ, $\Om_{i,k}=\{v: y_i(v)=k\}$, while $p_g(0\mid v)\ge\gamma$. Every $v\in\Om_{i,k}$ then receives the target $0$ with $m(v)=1$, and the cross-entropy gradient on the organ logit is
\begin{equation}
  \frac{\partial}{\partial z_{\theta,k}(v)}\Big[-\log p_\theta(0\mid v)\Big]=p_\theta(k\mid v)\;\ge 0 .
  \label{eq:extinction_grad}
\end{equation}
Gradient descent lowers $z_{\theta,k}(v)$ on the organ's own voxels until $p_\theta(k\mid v)\to 0$. When $g$ is tied to the student, the weakened student can yield a weaker teacher, and the organ is missed in further scans. This \emph{pseudo-label extinction} cannot be prevented by raising $\gamma$, because the offending targets are confident, nor by re-weighting organ classes, because no organ target exists on these voxels.

\section{Method}
\label{sec:method}

\begin{figure*}[t]
  \centering
  \includegraphics[width=0.9\textwidth]{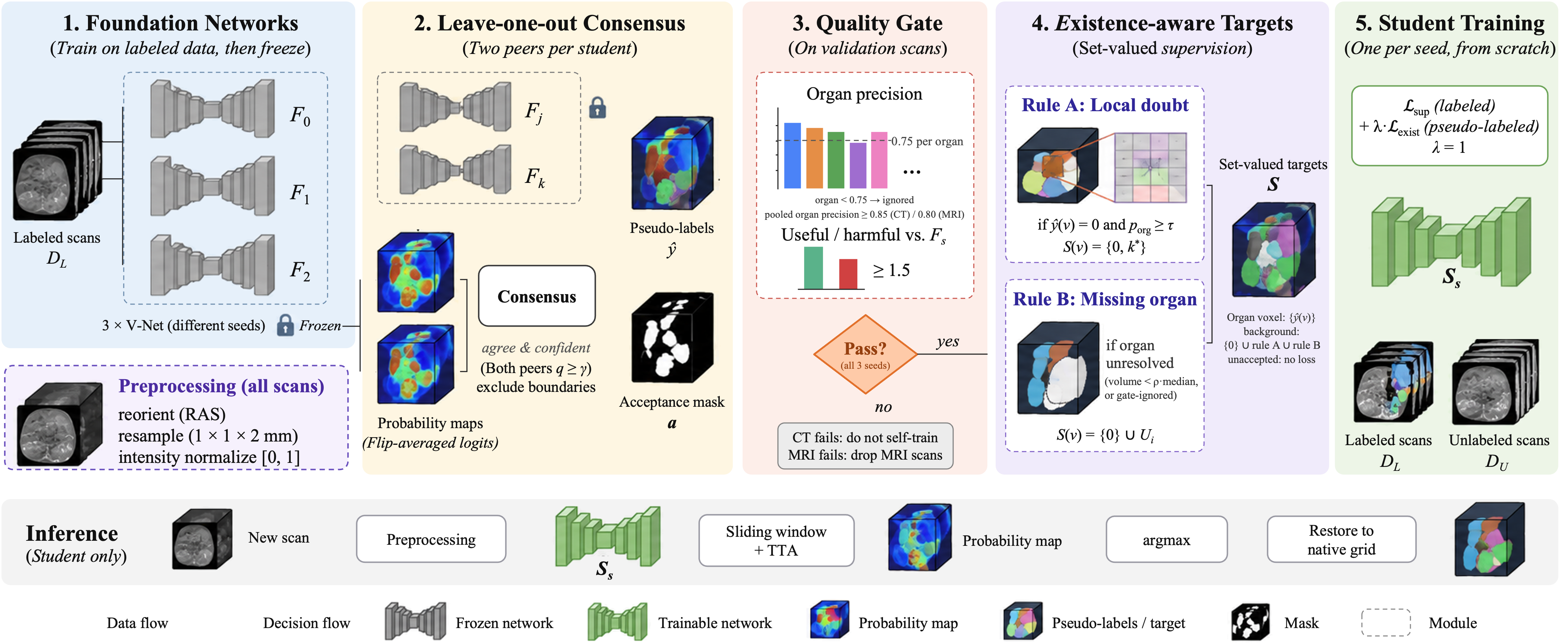}
  \caption{Overview of \method. Frozen foundation networks produce leave-one-out consensus pseudo-labels, a quality gate screens them, and the local doubt (Rule A) and missing-organ (Rule B) relaxations turn accepted background into set-valued targets for a trained student.}
  \label{fig:overview}
\end{figure*}

\label{sec:overview}
\Cref{fig:overview} shows \method, which has five stages. All scans are first reoriented to RAS, resampled to $1\times1\times2$\,mm, and intensity-normalized to $[0,1]$. (1)~Three foundation networks $F_0,F_1,F_2$ are trained on $\Dl$ only, with seeds $0,1,2$, and then frozen. (2)~For each student index $s\in\{0,1,2\}$, the two peers $\{F_j,F_k\}=\{F_0,F_1,F_2\}\setminus\{F_s\}$ label every unlabeled scan, and a voxel is accepted only where both agree with high confidence (\cref{sec:consensus}). (3)~A quality gate measured on $\Dv$ decides whether these pseudo-labels may be used and which organs are unreliable (\cref{sec:gate}). (4)~Accepted background is relaxed into set-valued targets wherever an organ may have been missed (\cref{sec:targets}). (5)~A student $S_s$ with the architecture of the foundation networks is trained from scratch on $\Dl\cup\Du$ (\cref{sec:loss}). Inference uses $S_s$ alone.

\subsection{Leave-one-out consensus pseudo-labels}
\label{sec:consensus}
For scan $x_i$ and peer $F_j$, we average the prediction over the identity, the anterior-posterior flip, the superior-inferior flip, and their combination, and denote the resulting posterior by $P^{j}_i(c\mid v)$. The left-right flip is not used because it would exchange the left and right kidneys and adrenal glands. Let $c_j(v)=\argmax_c P^{j}_i(c\mid v)$ be the label predicted by peer $j$ and $q_j(v)=P^{j}_i(c_j(v)\mid v)$ its confidence. The consensus label and the acceptance mask are
\begin{equation}
\begin{aligned}
  \hat y_i(v)&=c_j(v),\\
  a_i(v)&=\ind{c_j(v)=c_k(v)}\cdot\ind{v\notin\partial\hat y_i}\\
        &\quad\cdot\ind{\min(q_j(v),q_k(v))\ge\gamma},
\end{aligned}
  \label{eq:accept}
\end{equation}
with $\gamma=0.90$, where $\partial\hat y_i$ is the one-voxel band on each side of every class boundary of the consensus map. The accepted set is $\cA_i=\{v:a_i(v)=1\}$, with organ parts $\cA_i^{k}=\{v\in\cA_i:\hat y_i(v)=k\}$ and background part $\cB_i=\cA_i^{0}$. All other voxels are ignored and receive no loss. We also keep the peer-maximum posterior $\bar p_i(c\mid v)=\max\{P^{j}_i(c\mid v),P^{k}_i(c\mid v)\}$, which records whether at least one peer sees some evidence for class $c$ and is used in \cref{sec:targets}. Requiring the agreement of two independently trained networks is a form of disagreement-based uncertainty~\cite{lakshminarayanan2017simple}.

\subsection{Quality gate}
\label{sec:gate}
Pseudo-labels are useful only if their accepted organ voxels are mostly correct and if they add more correct information than they remove. We measure both properties by applying the pseudo-labeling of \cref{sec:consensus} to $\Dv$. The precision of the accepted voxels of organ $k$ is
\begin{equation}
  \mathrm{prec}_k=\Big(\sum_{i\in\Dv}\sum_{v\in\cA_i^{k}}\ind{y_i(v)=k}\Big)\Big/\sum_{i\in\Dv}|\cA_i^{k}|,
  \label{eq:precision}
\end{equation}
and the pooled precision is the same ratio computed over the accepted voxels of all organs. We further count an accepted voxel as \emph{useful} if its pseudo-label is correct while the held-out network $F_s$ is wrong, and as \emph{harmful} in the opposite case. The gate passes if the pooled precision is at least $0.85$ on CT ($0.80$ on MRI) and the useful-to-harmful ratio is at least $1.5$. If the gate fails on CT, self-training is not started, and if it fails only on MRI, the MRI scans are removed from $\Du$. Independently of this decision, any organ with $\mathrm{prec}_k<0.75$ is \emph{gate-ignored}: its accepted voxels are moved to the ignored set, and the organ is treated as unresolved in every scan.

\subsection{Existence-aware targets}
\label{sec:targets}
We replace the single label of an accepted voxel by a \emph{candidate set} $\cS_i(v)\subseteq\cC$ of classes that the student may predict without penalty. Accepted organ voxels keep a singleton, $\cS_i(v)=\{\hat y_i(v)\}$ for $v\in\cA_i^{k}$ and $k\in\cK$. Accepted background voxels start from $\cS_i(v)=\{0\}$ and are enlarged by two relaxations. The first is local (Rule~A in \cref{fig:overview}), and the second acts on a whole scan (Rule~B).

\paragraph{Local Doubt Relaxation (LDR).}
Let $k^\ast_i(v)=\argmax_{k\in\cK}\bar p_i(k\mid v)$ be the most likely organ at $v$ under the peers. If its probability is small but not negligible, the voxel may belong to a missed structure, and we admit the organ as a candidate:
\begin{equation}
  \cS_i(v)\leftarrow\{0,\,k^\ast_i(v)\}\quad\text{if } v\in\cB_i\ \text{and}\ \bar p_i\big(k^\ast_i(v)\mid v\big)\ge\tau,
  \label{eq:ruleA}
\end{equation}
with $\tau=0.02$, far below $\gamma$. LDR does not label the voxel as organ. It only withdraws the claim that the organ is absent.

\paragraph{Missing-Organ Relaxation (MOR).}
LDR cannot help when an organ is missed entirely, because the peers then assign it a negligible probability everywhere. We therefore test, per scan and per organ, whether the organ was found at all. Let $V_{i,k}=\nu\,|\cA_i^{k}|$ be the accepted volume of organ $k$ in scan $i$, with $\nu$ the voxel volume, and let $m_k$ be the median volume of organ $k$ over $\Dl$. Organ $k$ is \emph{unresolved} in scan $i$ if
\begin{equation}
  u_{i,k}=\ind{k\in\cK_{\mathrm{gate}}}\;\lor\;\ind{V_{i,k}<\rho\,m_k},\qquad \rho=0.25,
  \label{eq:unresolved}
\end{equation}
where $\cK_{\mathrm{gate}}$ is the set of gate-ignored organs. Since $V_{i,k}$ counts only voxels that both peers confidently assign to the organ, an organ that is under-segmented by more than a factor of four is also unresolved. With $\cU_i=\{k:u_{i,k}=1\}$, MOR enlarges every accepted background voxel of the scan:
\begin{equation}
  \cS_i(v)\leftarrow\cS_i(v)\cup\cU_i\qquad\text{for all } v\in\cB_i .
  \label{eq:ruleB}
\end{equation}
After both relaxations, a background voxel carries $\{0\}$ where no organ is in doubt, $\{0,k\}$ where one organ is in doubt, and a larger set where several organs are unresolved. The target map $\tilde Y_s=\{\cS_i(v)\}$ is computed once per student. MOR is global within a scan, because an organ that was not found could be anywhere.

\subsection{Learning objective}
\label{sec:loss}
A candidate set is trained with the negative log-likelihood of the event that the label lies in the set,
\begin{equation}
  \Lmce=-\frac{1}{|\cA_i|}\sum_{v\in\cA_i}\log\sum_{c\in\cS_i(v)}p_\theta(c\mid v),
  \label{eq:mce}
\end{equation}
which is the marginal loss of partial-label learning~\cite{jin2002learning,cour2011learning} and of partially supervised segmentation~\cite{shi2021marginal}. For a singleton it is the ordinary cross-entropy, and for $\{0,k\}$ it accepts either answer. The Dice loss, in contrast, needs a binary reference per organ. For organ $k$ we define the voxels at which its membership is \emph{certain}, $\cC_{i,k}=\{v\in\cA_i:\ k\notin\cS_i(v)\ \text{or}\ \cS_i(v)=\{k\}\}$, and compute the Dice term of $k$ only over $\cC_{i,k}$ with reference $r_{i,k}(v)=\ind{\cS_i(v)=\{k\}}$:
\begin{equation}
  \Ldice^{\sim}=1-\frac{1}{K}\sum_{k\in\cK}\frac{2\sum_{v\in\cC_{i,k}}p_\theta(k\mid v)\,r_{i,k}(v)+\epsilon}{\sum_{v\in\cC_{i,k}}\big(p_\theta(k\mid v)+r_{i,k}(v)\big)+\epsilon}.
  \label{eq:dice_skip}
\end{equation}
A voxel at which $k$ is allowed but not certain is excluded from the Dice term of organ $k$. Otherwise the denominator would penalize any mass placed on a candidate organ and partially undo \cref{eq:mce}. Each iteration draws labeled and pseudo-labeled patches and minimizes
\begin{equation}
\begin{aligned}
  \mathcal{L}(\theta)&=\Lsup(x_l,y_l)+\lambda\,\Lexist(x_u,\tilde Y),\\
  \Lexist&=\Lmce+\Ldice^{\sim},
\end{aligned}
  \label{eq:total}
\end{equation}
with $\lambda=1$ and deep supervision on both terms at every decoder scale. $\Lmce$ uses the class weights $w_c$ on singleton voxels.

\paragraph{Analysis.}
Fix a voxel with logits $z\in\R^{K+1}$, posterior $p=\mathrm{softmax}(z)$, and candidate set $\emptyset\neq\cS\subsetneq\cC$, and write $P_\cS=\sum_{c\in\cS}p_c$ and $\ell_\cS(z)=-\log P_\cS$.
\begin{proposition}
\label{prop:grad}
For every class $c$, $\partial\ell_\cS/\partial z_c=-\,p_c\,(1-P_\cS)/P_\cS\le 0$ if $c\in\cS$, and $\partial\ell_\cS/\partial z_c=p_c\ge 0$ otherwise. Moreover, $\ell_\cS$ depends on $z$ only through $A=\log\sum_{c\in\cS}e^{z_c}$ and $B=\log\sum_{c\notin\cS}e^{z_c}$, so the conditional distribution $p(c\mid c\in\cS)$ is unconstrained by $\ell_\cS$.
\end{proposition}
\begin{proof}
Since $\ell_\cS=\log\sum_{c\in\cC}e^{z_c}-\log\sum_{c\in\cS}e^{z_c}=\log(e^{A}+e^{B})-A$, the second claim is immediate, and differentiating the two terms gives $p_c-\ind{c\in\cS}\,p_c/P_\cS$.
\end{proof}
The first part states that gradient descent never decreases the logit of a candidate organ, in contrast to \cref{eq:extinction_grad}. The second part states that the loss carries no information about \emph{which} member of the set is correct. The student is relieved of the claim that the organ is absent, but it is not told that the organ is present, and the division of the mass within the set is left to the labeled patches and the accepted organ voxels~\cite{cour2011learning,feng2020provably}. \Cref{eq:dice_skip} preserves both properties, because the Dice term of a candidate organ does not depend on the voxels at which it is only allowed.

\subsection{Training and inference}
\label{sec:training}
\Cref{alg:method} summarizes the procedure. Foundation networks and students share the architecture, the optimizer, and an organ-balanced sampler, which draws an organ uniformly from $\cK$ and centers a $96^3$ patch on a random voxel of that organ (on unlabeled scans, a random accepted voxel). Students are trained from scratch and selected by validation Dice. At inference the student is applied in a sliding window of $96^3$ with $50\%$ overlap and Gaussian blending, the predictions of the four flips are averaged, and the argmax is resampled to the native grid.

\begin{algorithm}[t]
\caption{\method training for student index $s$}
\label{alg:method}
\begin{algorithmic}[1]
\Require $\Dl$, $\Du$, $\Dv$; thresholds $\gamma,\tau,\rho$; weight $\lambda$
\State Train $F_0,F_1,F_2$ on $\Dl$ with $\Lsup$ (seeds $0,1,2$); freeze; $\{j,k\}\leftarrow\{0,1,2\}\setminus\{s\}$
\For{each scan $x_i\in\Du\cup\Dv$}
  \State Flip-averaged $P^{j}_i,P^{k}_i$; consensus $\hat y_i$, mask $a_i$ by \cref{eq:accept}; $\bar p_i$
\EndFor
\State Gate on $\Dv$: pooled precision, $\mathrm{prec}_k$ by \cref{eq:precision}, useful-to-harmful ratio; $\cK_{\mathrm{gate}}\leftarrow\{k:\mathrm{prec}_k<0.75\}$; \textbf{stop} if the gate fails
\For{each scan $x_i\in\Du$}
  \State $\cS_i(v)\leftarrow\{\hat y_i(v)\}$ on $\cA_i$; ignore voxels of gate-ignored organs
  \State LDR by \cref{eq:ruleA}; $\cU_i$ by \cref{eq:unresolved}; MOR by \cref{eq:ruleB}
\EndFor
\State Initialize student $S_s$ with seed $s$
\For{each iteration}
  \State Organ-balanced patches $(x_l,y_l)\sim\Dl$, $(x_u,\tilde Y)\sim\Du$; step on \cref{eq:total}
\EndFor
\State \Return epoch of $S_s$ with the best validation Dice
\end{algorithmic}
\end{algorithm}

\begin{table*}[t]
\centering
\caption{Comparison with state-of-the-art methods on Synapse (20\% labeled), mean$_{\pm\text{std}}$ over three runs. Best in bold, second best underlined. $^\dagger$Trained at $1\times1\times2$\,mm and scored on the DHC grid. Other rows as reported by DCA~\cite{liu2026dca}.}
\label{tab:synapse}
\setlength{\tabcolsep}{2.6pt}
\renewcommand{\arraystretch}{0.85}
\scriptsize
\begin{tabular}{cl|cc|ccccccccccccc}
\toprule
& \multirow{2}{*}{\textbf{Method}} & \textbf{Avg.} & \textbf{Avg.} & \multicolumn{13}{c}{\textbf{Dice (\%) of each class}} \\
& & \textbf{Dice}$\boldsymbol{\uparrow}$ & \textbf{ASD}$\boldsymbol{\downarrow}$ & \textbf{Sp} & \textbf{RK} & \textbf{LK} & \textbf{Ga} & \textbf{Es} & \textbf{Li} & \textbf{St} & \textbf{Ao} & \textbf{IVC} & \textbf{PSV} & \textbf{Pa} & \textbf{RAG} & \textbf{LAG} \\
\midrule
 & V-Net (fully supervised) \cite{milletari2016vnet} & 62.09$_{\pm1.2}$ & 10.28$_{\pm3.9}$ & 84.6 & 77.2 & 73.8 & 73.3 & 38.2 & 94.6 & 68.4 & 72.1 & 71.2 & 58.2 & 48.5 & 17.9 & 29.0 \\
\midrule
\multirow{9}{*}{\rotatebox{90}{General}} & UA-MT \cite{yu2019uamt} & 20.26$_{\pm2.2}$ & 71.67$_{\pm7.4}$ & 48.2 & 31.7 & 22.2 & 0.0 & 0.0 & 81.2 & 29.1 & 23.3 & 27.5 & 0.0 & 0.0 & 0.0 & 0.0 \\
 & URPC \cite{luo2021urpc} & 25.68$_{\pm5.1}$ & 72.74$_{\pm15.5}$ & 66.7 & 38.2 & 56.8 & 0.0 & 0.0 & 85.3 & 33.9 & 33.1 & 14.8 & 0.0 & 5.1 & 0.0 & 0.0 \\
 & CPS \cite{chen2021cps} & 33.55$_{\pm3.7}$ & 41.21$_{\pm9.1}$ & 62.8 & 55.2 & 45.4 & 35.9 & 0.0 & 91.1 & 31.3 & 41.9 & 49.2 & 8.8 & 14.5 & 0.0 & 0.0 \\
 & SS-Net \cite{wu2022ssnet} & 35.08$_{\pm2.8}$ & 50.81$_{\pm6.5}$ & 62.7 & 67.9 & 60.9 & 34.3 & 0.0 & 89.9 & 20.9 & 61.7 & 44.8 & 0.0 & 8.7 & 4.2 & 0.0 \\
 & Co-BioNet \cite{peiris2023cobionet} & 58.83$_{\pm2.7}$ & 7.50$_{\pm5.8}$ & 82.8 & 90.0 & 86.5 & 11.6 & 19.5 & 92.3 & 47.7 & 77.5 & 77.7 & 51.3 & 30.3 & 47.5 & 50.2 \\
 & BCP \cite{bai2023bcp} & 50.23$_{\pm1.5}$ & 18.45$_{\pm2.3}$ & 72.4 & 58.1 & 52.7 & 38.2 & 12.5 & 92.3 & 35.6 & 48.9 & 45.2 & 18.7 & 22.1 & 8.4 & 5.2 \\
 & MagicNet \cite{chen2023magicnet} & 60.57$_{\pm2.5}$ & 22.48$_{\pm6.3}$ & 82.5 & 91.0 & 89.5 & 11.2 & 0.0 & 89.4 & 62.7 & 77.6 & 79.0 & 66.1 & 47.3 & 36.8 & 54.3 \\
 & ABD \cite{chi2024abd} & 55.67$_{\pm1.2}$ & 15.32$_{\pm1.8}$ & 76.8 & 64.3 & 58.9 & 42.5 & 18.9 & 93.7 & 41.2 & 55.4 & 51.6 & 24.3 & 28.7 & 14.6 & 11.8 \\
 & $\beta$-FFT \cite{hu2025betafft} & 59.12$_{\pm0.9}$ & 12.18$_{\pm1.1}$ & 81.2 & 70.5 & 65.4 & 48.7 & 25.3 & \underline{95.1} & 47.8 & 62.1 & 58.3 & 30.9 & 35.2 & 21.5 & 18.4 \\
\midrule
\multirow{10}{*}{\rotatebox{90}{Imbalance}} & Adsh \cite{guo2022adsh} & 35.29$_{\pm0.5}$ & 39.61$_{\pm4.6}$ & 55.1 & 59.6 & 45.8 & 52.2 & 0.0 & 89.4 & 32.8 & 47.6 & 53.0 & 8.9 & 14.4 & 0.0 & 0.0 \\
 & CReST \cite{wei2021crest} & 38.33$_{\pm3.4}$ & 22.85$_{\pm9.0}$ & 62.1 & 64.7 & 53.8 & 43.8 & 8.1 & 85.9 & 27.2 & 54.4 & 47.7 & 14.4 & 13.0 & 18.7 & 4.6 \\
 & SimiS \cite{chen2022simis} & 40.07$_{\pm0.6}$ & 32.98$_{\pm0.5}$ & 62.3 & 69.4 & 50.7 & 61.4 & 0.0 & 87.0 & 33.0 & 59.0 & 57.2 & 29.2 & 11.8 & 0.0 & 0.0 \\
 & Basak \etal \cite{basak2022addressing} & 33.24$_{\pm0.6}$ & 43.78$_{\pm2.5}$ & 57.4 & 53.8 & 48.5 & 46.9 & 0.0 & 87.8 & 28.7 & 42.3 & 45.4 & 6.3 & 15.0 & 0.0 & 0.0 \\
 & CLD \cite{lin2022cld} & 41.07$_{\pm1.2}$ & 32.15$_{\pm3.3}$ & 62.0 & 66.0 & 59.3 & 61.5 & 0.0 & 89.0 & 31.7 & 62.8 & 49.4 & 28.6 & 18.5 & 0.0 & 5.0 \\
 & DHC \cite{wang2023dhc} & 48.61$_{\pm0.9}$ & 10.71$_{\pm2.6}$ & 62.8 & 69.5 & 59.2 & \underline{66.0} & 13.2 & 85.2 & 36.9 & 67.9 & 61.5 & 37.0 & 30.9 & 31.4 & 10.6 \\
 & A\&D \cite{wang2023generic} & 60.88$_{\pm0.7}$ & 2.52$_{\pm0.4}$ & \underline{85.2} & 66.9 & 67.0 & 52.7 & \underline{62.9} & 89.6 & 52.1 & 83.0 & 74.9 & 41.8 & 43.4 & 44.8 & 27.2 \\
 & GA \cite{qi2024ga} & 68.43$_{\pm0.5}$ & 3.11$_{\pm0.2}$ & 81.4 & \underline{92.4} & \underline{90.8} & 33.5 & 53.3 & 89.1 & 60.9 & 79.1 & 82.1 & 66.7 & 48.7 & 50.3 & \underline{61.4} \\
 & SKCDF \cite{zhang2025skcdf} & 64.27$_{\pm1.4}$ & \textbf{1.45$_{\pm0.1}$} & 79.5 & 72.1 & 67.6 & 59.8 & 60.7 & 93.3 & 61.7 & \underline{85.4} & 78.5 & 41.8 & 50.9 & 46.4 & 37.8 \\
 & DCA \cite{liu2026dca} & \underline{73.20$_{\pm0.3}$} & \underline{1.78$_{\pm0.1}$} & 82.9 & \textbf{92.8} & \textbf{91.4} & 64.0 & 61.4 & 92.5 & \underline{67.2} & 83.9 & \underline{82.7} & \underline{68.1} & \underline{52.2} & \underline{51.8} & \textbf{63.7} \\
\midrule
 & \textbf{\method$^\dagger$} & \textbf{77.13$_{\pm0.4}$} & 2.60$_{\pm0.2}$ & \textbf{92.7} & 76.7 & 75.7 & \textbf{83.6} & \textbf{74.1} & \textbf{95.3} & \textbf{69.9} & \textbf{89.1} & \textbf{86.6} & \textbf{69.4} & \textbf{55.6} & \textbf{70.4} & \textbf{63.7} \\
\bottomrule
\end{tabular}
\end{table*}

\begin{table*}[t]
\centering
\caption{Comparison with state-of-the-art methods on AMOS (5\% labeled). Best in bold, second best underlined. $^\dagger$Mean over three seeds (std 0.07 Dice, 0.35 ASD), trained at $1\times1\times2$\,mm and scored on the $80\times160\times160$ DHC grid. Other rows as reported by DCA~\cite{liu2026dca}.}
\label{tab:amos}
\setlength{\tabcolsep}{2.6pt}
\renewcommand{\arraystretch}{0.85}
\scriptsize
\begin{tabular}{cl|cc|ccccccccccccccc}
\toprule
& \multirow{2}{*}{\textbf{Method}} & \textbf{Avg.} & \textbf{Avg.} & \multicolumn{15}{c}{\textbf{Dice (\%) of each class}} \\
& & \textbf{Dice}$\boldsymbol{\uparrow}$ & \textbf{ASD}$\boldsymbol{\downarrow}$ & \textbf{Sp} & \textbf{RK} & \textbf{LK} & \textbf{Ga} & \textbf{Es} & \textbf{Li} & \textbf{St} & \textbf{Ao} & \textbf{IVC} & \textbf{Pa} & \textbf{RAG} & \textbf{LAG} & \textbf{Du} & \textbf{Bl} & \textbf{P/U} \\
\midrule
 & V-Net (fully supervised) \cite{milletari2016vnet} & 76.50 & 2.01 & 92.2 & 92.2 & 93.3 & 65.5 & 70.3 & 95.3 & 82.4 & 91.4 & 85.0 & 74.9 & 58.6 & 58.1 & 65.6 & 64.4 & 58.3 \\
\midrule
\multirow{9}{*}{\rotatebox{90}{General}} & UA-MT \cite{yu2019uamt} & 42.16 & 15.48 & 59.8 & 64.9 & 64.0 & 35.3 & 34.1 & 77.7 & 37.8 & 61.0 & 46.0 & 33.3 & 26.9 & 12.3 & 18.1 & 29.7 & 31.6 \\
 & URPC \cite{luo2021urpc} & 44.93 & 27.44 & 67.0 & 64.2 & 67.2 & 36.1 & 0.0 & 83.1 & 45.5 & 67.4 & 54.4 & 46.7 & 0.0 & 29.4 & 35.2 & 44.5 & 33.2 \\
 & CPS \cite{chen2021cps} & 41.08 & 20.37 & 56.1 & 60.3 & 59.4 & 33.3 & 25.4 & 73.8 & 32.4 & 65.7 & 52.1 & 31.1 & 25.5 & 6.2 & 18.4 & 40.7 & 35.8 \\
 & SS-Net \cite{wu2022ssnet} & 33.88 & 54.72 & 65.4 & 68.3 & 69.9 & 37.8 & 0.0 & 75.1 & 33.2 & 68.0 & 56.6 & 33.5 & 0.0 & 0.0 & 0.0 & 0.2 & 0.2 \\
 & Co-BioNet \cite{peiris2023cobionet} & 48.32 & 26.04 & 76.6 & 82.1 & 75.1 & 41.5 & 38.2 & 87.9 & 40.4 & 75.2 & 53.7 & 40.8 & 4.8 & 0.0 & 25.1 & 64.2 & 19.2 \\
 & BCP \cite{bai2023bcp} & 45.67 & 28.12 & 62.3 & 66.1 & 62.5 & 34.8 & 18.7 & 79.4 & 38.2 & 64.3 & 49.1 & 35.6 & 22.4 & 15.2 & 22.8 & 35.1 & 32.9 \\
 & MagicNet \cite{chen2023magicnet} & 54.08 & 29.03 & 80.0 & 84.5 & 86.1 & 47.9 & 0.0 & 85.1 & 50.7 & 81.7 & 69.3 & 57.2 & 46.0 & 0.0 & 40.8 & 62.9 & 19.2 \\
 & ABD \cite{chi2024abd} & 50.23 & 22.45 & 68.7 & 70.4 & 68.2 & 39.5 & 25.3 & 84.2 & 42.1 & 69.8 & 54.7 & 41.2 & 28.6 & 21.8 & 29.4 & 42.3 & 38.7 \\
 & $\beta$-FFT \cite{hu2025betafft} & 54.89 & 18.76 & 74.2 & 75.8 & 73.6 & 45.2 & 32.1 & 88.9 & 48.7 & 75.4 & 60.3 & 47.8 & 35.2 & 28.9 & 36.7 & 49.5 & 45.1 \\
\midrule
\multirow{10}{*}{\rotatebox{90}{Imbalance}} & Adsh \cite{guo2022adsh} & 40.33 & 24.53 & 56.0 & 63.6 & 57.3 & 34.7 & 25.7 & 73.9 & 30.7 & 65.7 & 51.9 & 27.1 & 20.2 & 0.0 & 18.6 & 43.5 & 35.9 \\
 & CReST \cite{wei2021crest} & 46.55 & 14.62 & 66.5 & 64.2 & 65.4 & 36.0 & 32.2 & 77.8 & 43.6 & 68.5 & 52.9 & 40.3 & 24.7 & 19.5 & 26.5 & 43.9 & 36.4 \\
 & SimiS \cite{chen2022simis} & 47.27 & 11.51 & 77.4 & 72.5 & 68.7 & 32.1 & 14.7 & 86.6 & 46.3 & 74.6 & 54.2 & 41.6 & 24.4 & 17.9 & 21.9 & 47.9 & 28.2 \\
 & Basak \etal \cite{basak2022addressing} & 38.74 & 31.76 & 68.8 & 59.0 & 54.2 & 29.0 & 0.0 & 83.7 & 39.3 & 61.7 & 52.1 & 34.6 & 0.0 & 0.0 & 26.8 & 45.7 & 26.2 \\
 & CLD \cite{lin2022cld} & 46.10 & 15.86 & 67.2 & 68.5 & 71.4 & 41.0 & 21.0 & 76.1 & 42.4 & 69.8 & 52.1 & 37.9 & 24.7 & 23.4 & 22.7 & 38.1 & 35.2 \\
 & DHC \cite{wang2023dhc} & 49.53 & 13.89 & 68.1 & 69.6 & 71.1 & 42.3 & 37.0 & 76.8 & 43.8 & 70.8 & 57.4 & 43.2 & 27.0 & 28.7 & 29.1 & 41.4 & 36.7 \\
 & A\&D \cite{wang2023generic} & 50.03 & 5.21 & 73.1 & 76.0 & 76.5 & 29.1 & 44.9 & 82.5 & 49.0 & 72.8 & 61.7 & 48.5 & 30.2 & 19.7 & 36.4 & 32.9 & 18.2 \\
 & GA \cite{qi2024ga} & 63.51 & 4.58 & 78.9 & 85.5 & 87.2 & 50.0 & 49.1 & 86.9 & 56.2 & 83.4 & 70.3 & 57.4 & 49.1 & 40.8 & 38.3 & \underline{71.6} & \underline{47.9} \\
 & SKCDF \cite{zhang2025skcdf} & 53.81 & 5.97 & 77.1 & 77.9 & 71.2 & 34.1 & 50.4 & 88.6 & 51.6 & 80.9 & 58.9 & 48.8 & 33.0 & 30.2 & 32.2 & 45.9 & 26.4 \\
 & DCA \cite{liu2026dca} & \underline{69.90} & \textbf{2.66} & \underline{87.3} & \underline{87.8} & \underline{90.6} & \underline{59.6} & \underline{63.6} & \underline{91.3} & \underline{66.4} & \underline{87.0} & \underline{76.7} & \underline{64.0} & \underline{55.8} & \underline{48.5} & \underline{46.1} & \textbf{72.7} & \textbf{51.1} \\
\midrule
 & \textbf{\method$^\dagger$} & \textbf{75.91} & \underline{3.40} & \textbf{93.1} & \textbf{93.0} & \textbf{94.6} & \textbf{64.3} & \textbf{75.6} & \textbf{94.9} & \textbf{73.7} & \textbf{92.2} & \textbf{87.4} & \textbf{78.0} & \textbf{67.9} & \textbf{66.0} & \textbf{63.3} & 54.3 & 40.4 \\
\bottomrule
\end{tabular}
\end{table*}

\section{Experiments}
\label{sec:experiments}

\paragraph{Datasets and protocol.}
We follow the class-imbalanced protocol of DHC~\cite{wang2023dhc}, which is also used by A\&D, GA, SKCDF, and DCA, so that all numbers in \cref{tab:synapse,tab:amos} share splits, label budgets, and evaluator.
\textbf{Synapse}~\cite{landman2015btcv} (BTCV) has 30 abdominal CT scans with 13 organs: spleen (Sp), right and left kidney (RK, LK), gallbladder (Ga), esophagus (Es), liver (Li), stomach (St), aorta (Ao), inferior vena cava (IVC), portal and splenic veins (PSV), pancreas (Pa), and right and left adrenal gland (RAG, LAG). The split has 20 training, 4 validation, and 6 test scans, and 20\% of the training scans (4) are labeled. We use the split released with SKCDF~\cite{zhang2025skcdf}.
\textbf{AMOS}~\cite{ji2022amos} has 360 CT and MRI scans with 15 organs: those above without PSV, plus duodenum (Du), bladder (Bl), and prostate/uterus (P/U). The DHC split has 216 training, 24 validation, and 120 test scans, with 5\% of the training scans labeled. Dice (\%) and average surface distance (ASD, in voxels of the evaluation grid) are computed per organ with the DHC evaluator and averaged over organs and test scans. \method is trained and applied at $1\times1\times2$\,mm, and its predictions are resampled to the evaluation grid (in-plane halved for Synapse, $80\times160\times160$ for AMOS) before scoring. Organs that are not annotated in a scan (bladder and prostate/uterus in the MRI scans of AMOS) are scored against an empty reference, as in the evaluator.

\paragraph{Baselines.}
We compare against the 19 semi-supervised methods reported by DCA~\cite{liu2026dca} under this protocol, nine general and ten class-imbalance methods, and against a V-Net trained on all labels with the recipe of the protocol. These numbers are taken from DCA, where each method uses its own recipe and input resolution. Since \method uses a residual V-Net at $1\times1\times2$\,mm, row~(a) of \cref{tab:ablation} reports our network trained on the labeled scans alone, which isolates the contribution of backbone and resolution. For \cref{fig:qual_synapse,fig:tsne_synapse} we ran DHC, SKCDF, GA, and MagicNet with released code and BCP through our multi-organ 3D port.

\paragraph{Implementation.}
All networks are residual V-Nets~\cite{milletari2016vnet} with five levels of 32 to 512 channels, instance normalization, and deep supervision at four decoder scales. CT intensities are clipped to $[-75,275]$\,HU and MRI intensities to the 0.5 to 99.5 percentile range before scaling to $[0,1]$. Foundation networks and students are trained for 40k iterations with AdamW (learning rate $3\times10^{-4}$, weight decay $10^{-4}$, warm-up followed by cosine decay) on batches of two labeled and, for students, two pseudo-labeled $96^3$ patches. Augmentation consists of the two flips of \cref{sec:consensus}, intensity perturbations, and mild rotation and scaling. The class weights are $w_c\propto\sqrt{N_{\mathrm{fg}}/N_c}$ for organs, with $N_c$ the voxel count of class $c$ in $\Dl$, and $0.1$ for background. We use $\gamma=0.90$, $\tau=0.02$, $\rho=0.25$, $\lambda=1$, and the gate thresholds of \cref{sec:gate} on both benchmarks. Results are the mean and standard deviation over the three students (seeds 0, 1, 2). The network has 31.0M parameters. On one NVIDIA A100, the pipeline for one student, including the three foundation networks, takes 11.2 GPU hours on Synapse and 31.1 on AMOS, against 2.3 and 5.4 for the labeled-only network. Inference takes 15.1 and 18.7 seconds per scan in both cases.

\section{Results}
\label{sec:results}

\subsection{Comparison with the state of the art}
\label{sec:sota}
\paragraph{Synapse.}
With 20\% labeled scans (\cref{tab:synapse}), \method reaches 77.13 mean Dice, against 73.20 for DCA, 68.43 for GA, and 64.27 for SKCDF. It obtains the best Dice on 11 of the 13 organs, including a tie with DCA on the left adrenal gland. The largest gains over DCA are on small organs: 19.6 Dice on the gallbladder (83.6 against 64.0), 18.6 on the right adrenal gland (70.4 against 51.8), and 12.7 on the esophagus (74.1 against 61.4). The two kidneys are the exception: \method reaches 76.7 and 75.7 Dice, close to the fully supervised V-Net (77.2 and 73.8) but well below DCA (92.8 and 91.4).

\paragraph{AMOS.}
With 5\% labeled scans (\cref{tab:amos}), \method reaches 75.91 mean Dice, against 69.90 for DCA and 63.51 for GA, which is within 0.6 Dice of the fully supervised V-Net (76.50). \method is the best method on 13 of the 15 organs, and the gains over DCA again grow as the organs become smaller: 17.5 Dice on the left adrenal gland, 17.2 on the duodenum, 14.0 on the pancreas, 12.1 on the right adrenal gland, and 12.0 on the esophagus. Its ASD (3.40) is the second lowest after DCA (2.66). DCA has released neither code nor per-case results, which rules out a paired test. Against DHC, re-run with its released code, \method is higher on all 120 test scans (mean difference 24.0 Dice, Wilcoxon signed-rank $p=2.0\times10^{-21}$). The bladder (54.3 against 72.7 for DCA) and the prostate/uterus (40.4 against 51.1) are the two organs on which \method is worse. The bladder is the one organ that the quality gate ignores on AMOS, so it never receives an organ pseudo-label and is treated as unresolved in every scan. Both organs are also not annotated in the MRI scans, where MOR still permits the student to predict them.

\paragraph{Organ size.}
\Cref{fig:per_organ} relates these results to organ size, with the organs sorted by their share of the labeled foreground. On the right of the plot, where the share is below one percent, the curve of \method lies at or above those of DCA, GA, SKCDF, and DHC, with the prostate/uterus on AMOS as the only exception. The baselines drop sharply there, whereas \method stays between 63 and 84 Dice on Synapse and above 65 Dice on the esophagus and adrenal glands of AMOS. Part of the margin over the published numbers comes from the backbone and resolution.

\begin{figure*}[t]
  \centering
  \includegraphics[width=0.8\textwidth]{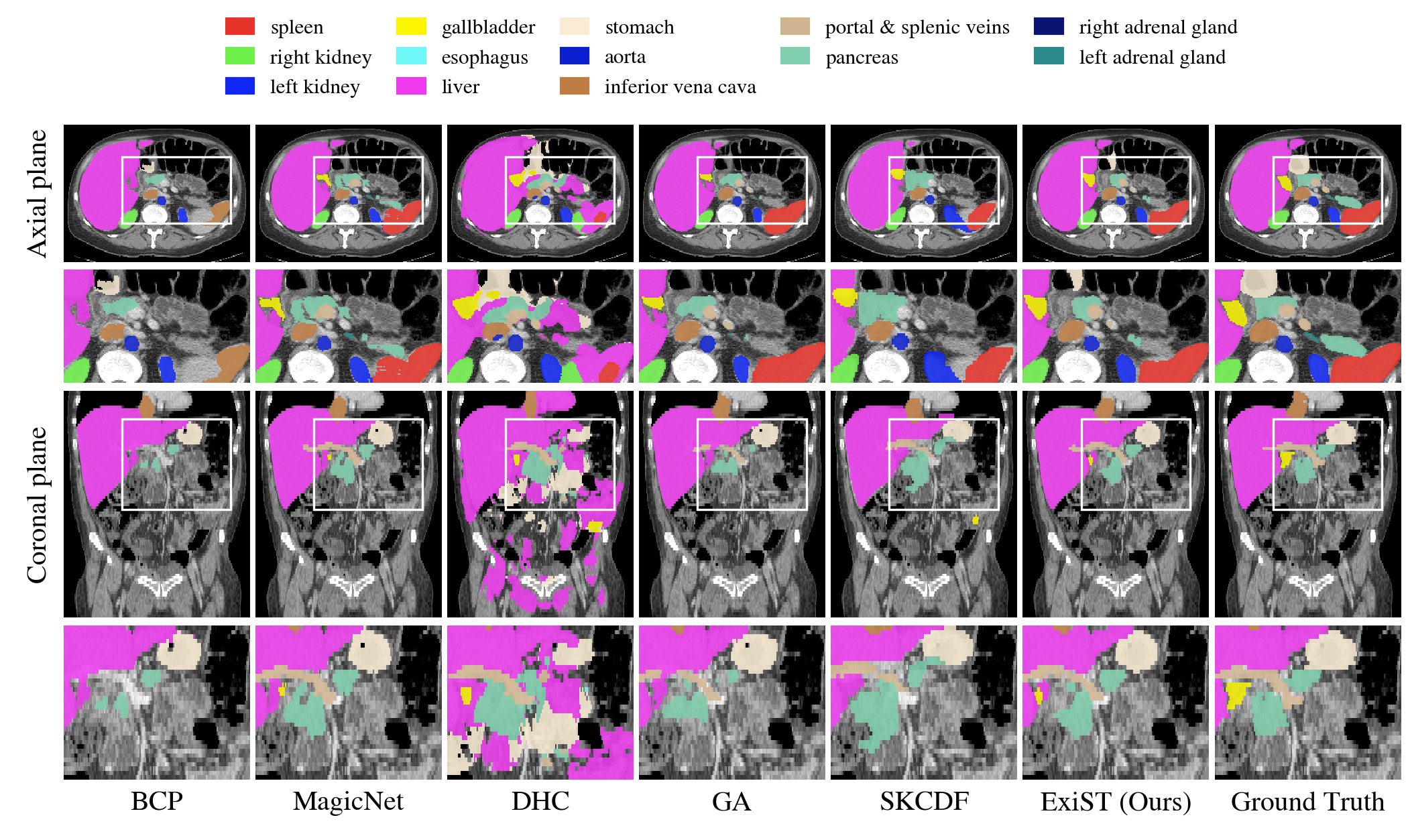}
  \caption{Qualitative comparison on the Synapse test scan with the median Dice of \method (20\% labeled). The slices pass through the reference pancreas, and the white boxes mark the enlarged regions.}
  \label{fig:qual_synapse}
\end{figure*}

\subsection{Qualitative comparison}
\label{sec:qualitative}
\Cref{fig:qual_synapse} shows the Synapse test scan with the median Dice of \method, fixed before inspection. In the axial view, \method segments the gallbladder with a shape close to the reference. BCP misses it, GA and MagicNet recover only a fragment, and DHC produces scattered false positives of several organs. In the coronal view, the portal and splenic veins and the two visible parts of the pancreas predicted by \method follow the reference, whereas BCP and GA segment the pancreas only partially and SKCDF extends it well beyond the reference. The remaining errors of \method are also visible: the stomach is under-segmented in the axial slice, and only a fragment of the pancreatic tail is found.

\begin{figure}[t]
  \centering
  \includegraphics[width=0.8\linewidth]{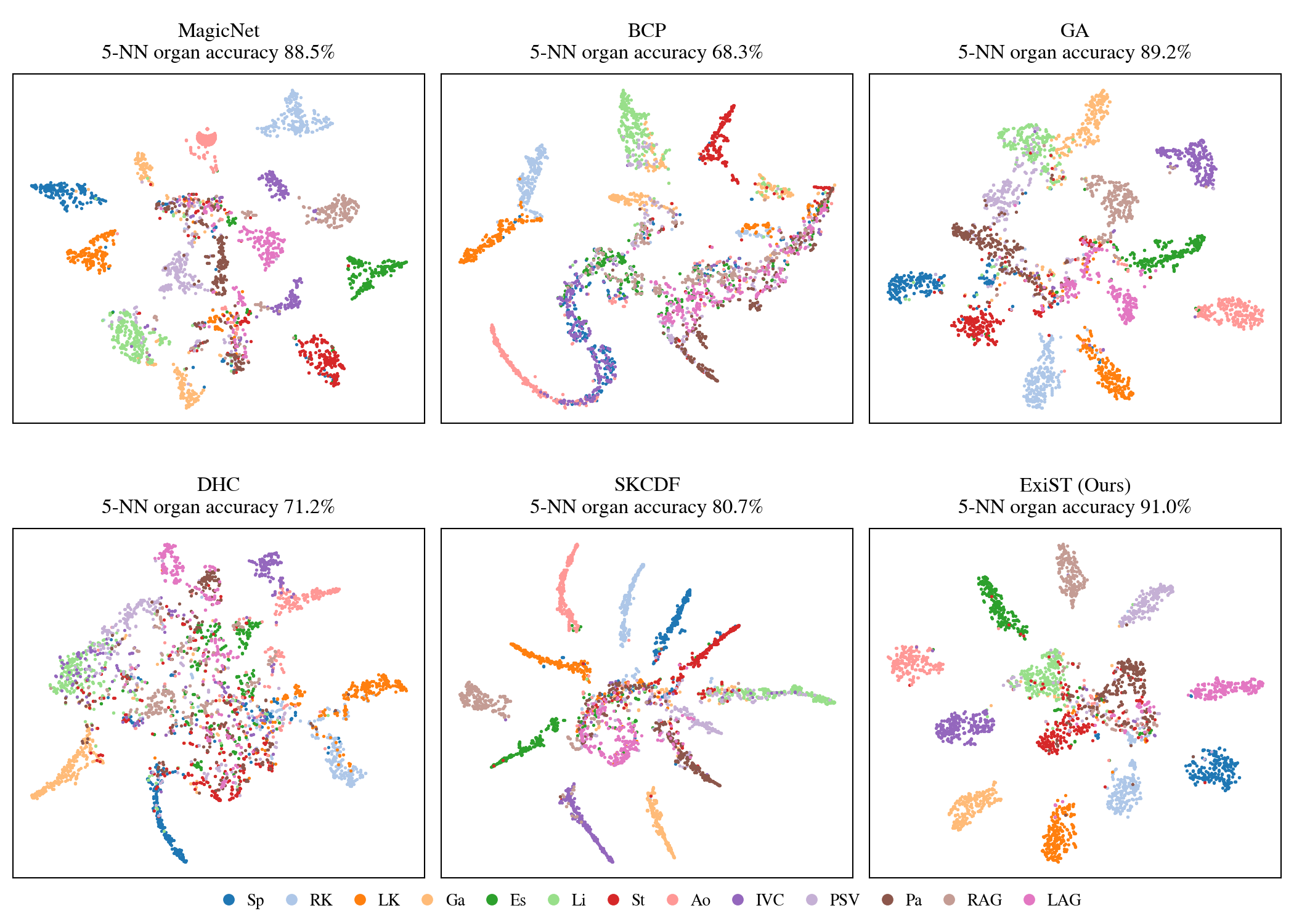}
  \caption{t-SNE of per-voxel features before the classifier on the Synapse test set (40 voxels per organ and scan).}
  \label{fig:tsne_synapse}
\end{figure}

\subsection{Feature-space analysis}
\label{sec:features}
\Cref{fig:tsne_synapse} visualizes the per-voxel features of the penultimate layer on the Synapse test set. For \method, most organs form compact and separate clusters, including the gallbladder, the esophagus, and the left adrenal gland. The pancreas, the stomach, and part of the right adrenal gland overlap in the center of the plot, which matches their lower Dice in \cref{tab:synapse}. The features of DHC and BCP mix most organs in one central cloud. The leave-one-out 5-nearest-neighbor organ accuracy in the original feature space summarizes this impression: 91.0\% for \method, against 89.2\% for GA, 88.5\% for MagicNet, 80.7\% for SKCDF, 71.2\% for DHC, and 68.3\% for BCP. Part of the difference may come from the input resolution of each method.

\section{Ablation Studies}
\label{sec:ablation}

\begin{table}[t]
  \centering
  \caption{Component ablation on Synapse (20\% labeled). Tail is the mean Dice over Es, Pa, RAG, and LAG. $^\dagger$Earlier gate configuration, under which the gallbladder is gate-ignored.}
  \label{tab:ablation}
  \footnotesize
  \setlength{\tabcolsep}{3pt}
  \begin{tabular}{@{}lccc@{}}
    \toprule
    \textbf{Variant} & \textbf{Dice} & \textbf{ASD} & \textbf{Tail} \\
    \midrule
    (a) Foundation only, no self-training & 74.23$_{\pm0.7}$ & 2.70$_{\pm0.2}$ & 63.1 \\
    (b) (a) + consensus, no relaxation & 75.20$_{\pm0.6}$ & 2.76$_{\pm0.0}$ & 63.9 \\
    \midrule
    (c) (b) + LDR only$^\dagger$ & 75.77$_{\pm0.5}$ & 2.68$_{\pm0.2}$ & 64.8 \\
    (d) (b) + MOR only$^\dagger$ & 72.83$_{\pm0.9}$ & 6.71$_{\pm0.8}$ & 66.2 \\
    (e) (b) + LDR + MOR, standard Dice$^\dagger$ & 73.53$_{\pm1.0}$ & 3.05$_{\pm0.7}$ & 63.8 \\
    (f) Full$^\dagger$ & 73.67$_{\pm0.6}$ & 5.79$_{\pm1.3}$ & 66.3 \\
    (g) Full, peers include $F_s$$^\dagger$ & 73.81$_{\pm0.5}$ & 4.90$_{\pm0.7}$ & \textbf{66.9} \\
    (h) Full, single peer$^\dagger$ & 71.98$_{\pm0.8}$ & 6.67$_{\pm2.2}$ & 64.4 \\
    \midrule
    \textbf{Full \method} & \textbf{77.13$_{\pm0.4}$} & \textbf{2.60$_{\pm0.2}$} & 65.9 \\
    \bottomrule
  \end{tabular}
\end{table}

\paragraph{Effectiveness of key components.}
\Cref{tab:ablation} separates the ingredients of \method on Synapse. Row~(a) is the foundation network trained on the four labeled scans alone. It reaches 74.23 Dice, which is already above the published methods of \cref{tab:synapse}, so the following rows must be read against this strong baseline. Self-training with consensus pseudo-labels and no relaxation (b) adds 1.0 Dice and 0.8 tail Dice. The full method adds another 1.9 Dice and 2.0 tail Dice and reaches the lowest ASD of the table (2.60). In total, \method gains 2.9 Dice and 2.8 tail Dice over its labeled-only baseline.

Rows (c) to (h) were run with an earlier configuration of the gate, under which the gallbladder was gate-ignored on Synapse, and are compared with row~(f), the full method under that configuration. Among them, LDR alone (c) has the highest mean Dice (75.77) and the lowest ASD (2.68). MOR alone (d) reaches a higher tail Dice (66.2 against 64.8) but a lower mean Dice (72.83) and an ASD of 6.71, so under this configuration it recovers small organs at the price of false positives. Replacing the uncertainty-skipping Dice by the standard Dice (e) removes this price (ASD 3.05) but also the gain on tail organs (63.8 against 66.3 in f), which confirms that a standard Dice term penalizes the mass that the marginal loss permits. The agreement of two peers matters, since a single peer (h) loses 1.7 Dice and 1.9 tail Dice relative to (f). Including the own-seed network $F_s$ among the peers (g) changes the result by less than one standard deviation, so the leave-one-out exclusion is a safeguard rather than a source of the gain.

\paragraph{Gate-ignored organs.}
Row~(f) and the full method differ only in the gate configuration, which changes mean Dice from 73.67 to 77.13 and ASD from 5.79 to 2.60. With the gallbladder gate-ignored, its accepted voxels are discarded and MOR is applied to it in every scan, so the organ loses all positive pseudo-labels while every background voxel may become gallbladder. Relaxation thus helps for organs that the peers actually miss, whereas a blanket relaxation of an organ that they mostly find is harmful, as for the bladder on AMOS (\cref{tab:amos}). The gate ratio is also informative across seeds. On Synapse the pooled precision is at least 0.97 for all three peer pairs, and the useful-to-harmful ratios are 0.77, 7.09, and 10.04. The first value is below the threshold of 1.5, but we trained this student nevertheless, and it is the weakest of the three (76.72 Dice against 76.94 and 77.73).

\begin{table}[t]
  \centering
  \caption{Pseudo-labeler design on AMOS. $\Delta$Dice is teacher minus student, and Ga Dice is given before and after self-training.}
  \label{tab:pilot_main}
  \footnotesize
  \setlength{\tabcolsep}{3pt}
  \begin{tabular}{@{}lccc@{}}
    \toprule
    \textbf{Pseudo-labeler} & \textbf{$\Delta$Dice} & \textbf{Useful:harm.} & \textbf{Ga Dice} \\
    \midrule
    EMA teacher, hard labels & $+0.07$ & $3.76$ & $65.8\rightarrow66.1$ \\
    Consensus + LDR + MOR (ours) & n/a & $2.21$ & $65.8\rightarrow71.3$ \\
    \bottomrule
  \end{tabular}
\end{table}

\paragraph{Pseudo-labeler design.}
\Cref{tab:pilot_main} compares the frozen peers with a single EMA teacher that is trained with our recipe and hard pseudo-labels on AMOS. The EMA teacher is only 0.07 Dice better than its student, so its pseudo-labels carry little information that the student does not already have. Its useful-to-harmful ratio is higher than that of the consensus (3.76 vs. 2.21), but the gallbladder gains only 0.3 Dice during self-training, against 5.5 Dice with \method. Under our recipe the EMA teacher does not extinguish the gallbladder, because the labeled-only network already finds it in most scans.

\paragraph{Relaxation thresholds.}
We varied one threshold at a time on Synapse in the configuration of row~(f) of \cref{tab:ablation} (73.67 Dice at the defaults). The acceptance confidence $\gamma=0.80$ and $0.95$ gives 72.44 and 73.57, the doubt threshold $\tau=0.01$, $0.05$, and $0.10$ gives 73.24, 73.83, and 72.83, and the existence ratio $\rho=0$, $0.10$, and $0.50$ gives 73.34, 72.44, and 73.41. All settings lie within 1.4 Dice, about twice the std. over seeds, so the method is not sensitive to its thresholds.

\section{Conclusion}
\label{sec:conclusion}
Standard self-training tells the student that a missed organ is absent, which extinguishes small organs. \method replaces this claim by \textit{unresolved} through gated consensus pseudo-labels, a per-scan and per-organ existence test, and set-valued targets whose marginal loss never pushes a candidate organ toward absence. On Synapse and AMOS, \method improves mean Dice by 3.9 and 6.0 points, with the largest gains on small organs and single-network inference. Since the relaxation hurts when applied to an organ in every scan, future existence tests should also use anatomy and modality.

\section*{LLM Usage Statement}
Generative AI tools were used for language editing of the manuscript and to assist with the literature search. They were not used to generate experimental results, numerical values, or the statements and proofs of the theoretical results, all of which were derived by the authors. The authors take full responsibility for the final content of this work, including all text, claims and artifacts.
{
    \small
    \bibliographystyle{ieeenat_fullname}
    \bibliography{main}
}


\end{document}